\documentclass[acmtog]{acmart}
\acmSubmissionID{539}

\usepackage{booktabs} 
\usepackage{multirow} 
\usepackage{xcolor} 
\usepackage{colortbl} 
\usepackage{threeparttable} 
\usepackage{pdfpages}
\usepackage{subfigure} 
\usepackage{hyperref} 

\usepackage{amssymb, amsfonts} 
\usepackage{verbatim}
\usepackage{lipsum}
\newcommand{\BestCellColor}{\cellcolor{red!25}}
\newcommand{\SecondBestCellColor}{\cellcolor{orange!25}}

\definecolor{best1}{RGB}{222,242,212}
\definecolor{best2}{RGB}{255,250,212}
\makeatletter
\DeclareRobustCommand\onedot{\futurelet\@let@token\@onedot}
\def\@onedot{\ifx\@let@token.\else.\null\fi\xspace}

\makeatother

\usepackage[ruled]{algorithm2e} 

\SetAlFnt{\small}
\SetAlCapFnt{\small}
\SetAlCapNameFnt{\small}
\SetAlCapHSkip{0pt}

\setcopyright{rightsretained}
\acmJournal{TOG}
\acmYear{2024} \acmVolume{43} \acmNumber{6} \acmArticle{} \acmMonth{12}\acmDOI{10.1145/3687935}

\usepackage{array}
\newcolumntype{H}{>{\setbox0=\hbox\bgroup}c<{\egroup}@{}}

\begin{document}
\title{V\textsuperscript{3}: Viewing Volumetric Videos on Mobiles via Streamable 2D Dynamic Gaussians}

\author{Penghao Wang}\authornote{Equal contributions.}
\orcid{0009-0004-7440-0370}
\affiliation{%
 \institution{ShanghaiTech University and NeuDim Digital Technology (Shanghai) Co.,Ltd.}
 \city{Shanghai}
 \country{China}}
\email{wangph1@shanghaitech.edu.cn}

\author{ZhiRui Zhang}\authornotemark[1]
\orcid{0009-0008-8171-4721}
\affiliation{%
 \institution{ShanghaiTech University and NeuDim Digital Technology (Shanghai) Co.,Ltd.}
 \city{Shanghai}
 \country{China}}
\email{zhangzhr4@shanghaitech.edu.cn}

\author{Liao Wang}\authornotemark[1]
\orcid{0000-0002-6235-8628}
\affiliation{%
 \institution{ShanghaiTech University and NeuDim Digital Technology (Shanghai) Co.,Ltd.}
 \city{Shanghai}
 \country{China}}
\email{wangla@shanghaitech.edu.cn}

\author{Kaixin Yao}
\orcid{0009-0005-2056-6057}
\affiliation{%
 \institution{ShanghaiTech University and NeuDim Digital Technology (Shanghai) Co.,Ltd.}
 \city{Shanghai}
 \country{China}}
\email{yaokx2023@shanghaitech.edu.cn}

\author{Siyuan Xie}
\orcid{0009-0006-1158-9668}
\affiliation{%
 \institution{ShanghaiTech University and NeuDim Digital Technology (Shanghai) Co.,Ltd.}
 \city{Shanghai}
 \country{China}}
\email{xiesy2022@shanghaitech.edu.cn}

\author{Jingyi Yu}\authornotemark[2]
\orcid{0000-0002-8580-0036}
\affiliation{%
 \institution{ShanghaiTech University.}
 \city{Shanghai}
 \country{China}}
\email{yujingyi@shanghaitech.edu.cn}

\author{Minye Wu}\authornotemark[2]
\orcid{0000-0002-8163-9513}
\affiliation{%
 \institution{KU Leuven.}
 \city{Leuven}
 \country{Belgium}}
\email{minye.wu@kuleuven.be}

\author{Lan Xu}\authornote{Corresponding author.}
\orcid{0000-0002-8807-7787}
\affiliation{%
 \institution{ShanghaiTech University.}
 \city{Shanghai}
 \country{China}}
\email{xulan1@shanghaitech.edu.cn}

\begin{abstract}
%



Experiencing high-fidelity volumetric video as seamlessly as 2D videos is a long-held dream. However, current dynamic 3DGS methods, despite their high rendering quality, face challenges in streaming on mobile devices due to computational and bandwidth constraints.
In this paper, we introduce V\textsuperscript{3}(Viewing Volumetric Videos), a novel approach that enables high-quality mobile rendering through the streaming of dynamic Gaussians. Our key innovation is to view dynamic 3DGS as 2D 
videos, facilitating the use of hardware video codecs. Additionally, we propose a two-stage training strategy to reduce storage requirements with rapid training speed. The first stage employs hash encoding and shallow MLP to learn motion, then reduces the number of Gaussians through pruning to meet the streaming requirements, while the second stage fine tunes other Gaussian attributes using residual entropy loss and temporal loss to improve temporal continuity.
This strategy, which disentangles motion and appearance, maintains high rendering quality with compact storage requirements.
Meanwhile, we designed a multi-platform player to decode and render 2D Gaussian videos.
Extensive experiments demonstrate the effectiveness of V\textsuperscript{3}, outperforming other methods by enabling high-quality rendering and streaming on common devices, which is unseen before. 
As the first to stream dynamic Gaussians on mobile devices, our companion player offers users an unprecedented volumetric video experience, including smooth scrolling and instant sharing. Our project page with source code is available at \textcolor{blue}{\url{https://authoritywang.github.io/v3/}}.
\end{abstract}

\ccsdesc[500]{Computing methodologies~Computer graphics}

\keywords{Neural rendering, 3D Gaussian Splatting, human performance, volumetric video, mobile rendering}

\begin{teaserfigure}
 \centering
 \includegraphics[width=1.0\linewidth]{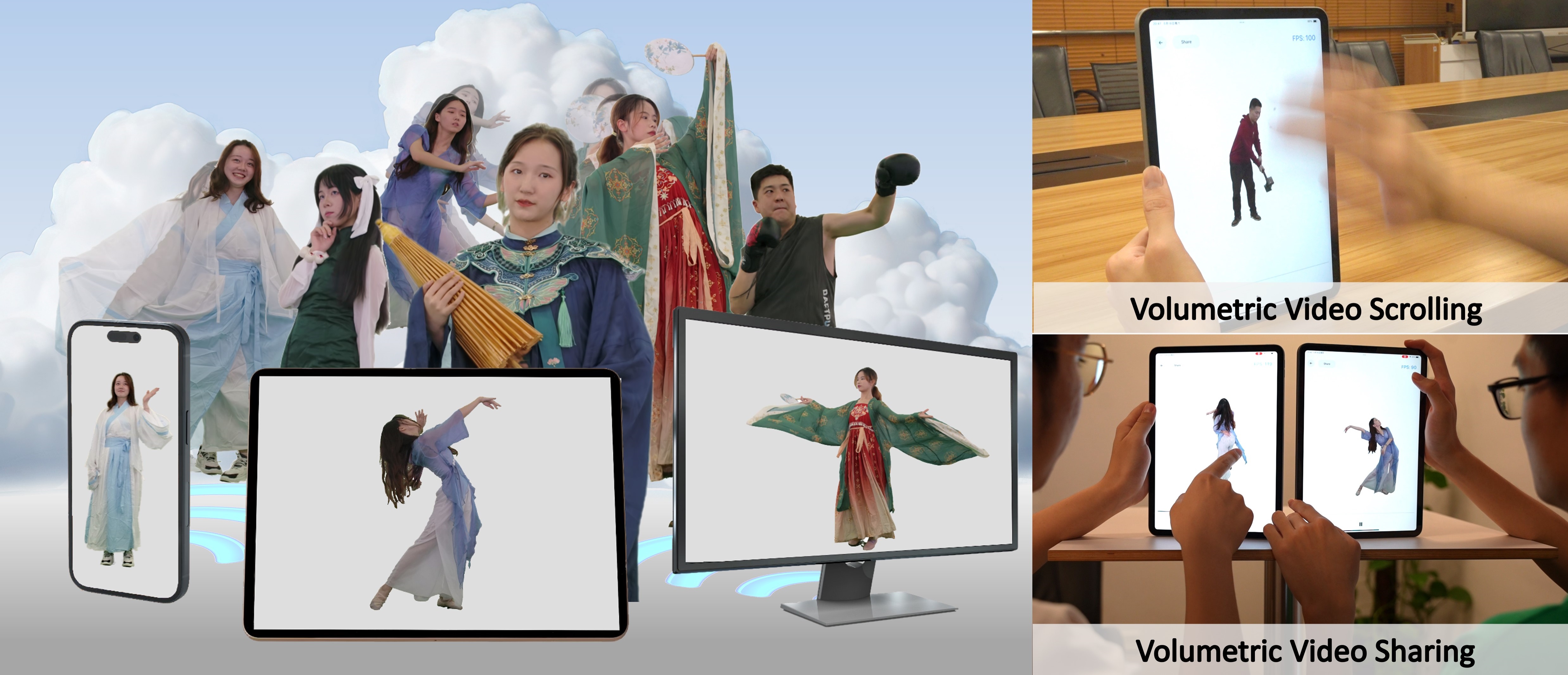}
 \vspace{-21pt}
 \caption{
Our method can stream 2D Gridded Gaussians to mobiles for high-quality rendering with low storage requirements, providing users with a unique volumetric video viewing experience across multiple devices.}
 \label{fig:teaser}
\end{teaserfigure}

\maketitle

\section{Introduction}
In the era of mobile internet, we humans can conveniently enjoy 2D videos on mobile devices anytime, anywhere. Further viewing photo-real volumetric videos on mobile devices, can allow users to freely choose their viewing angles and provide immersive experiences. However, streaming and rendering high-quality volumetric video to mobiles remains challenging. The difficulty lies in maintaining high fidelity with computational complexity suitable for mobile devices while supporting streaming with limited storage.

Traditional volumetric video reconstruction often relies on mesh reconstruction and texture mapping~\cite{newcombe2015dynamicfusion, zhao2022human}. Although dynamic meshes can support streaming on mobile devices, the precision of mesh-based methods is limited, especially in areas with occlusion, lack of texture, or parts like hair that are difficult to represent with meshes. 
Neural Radiance Field (NeRF) can avoid these issues caused by geometric reconstruction and achieve photorealistic effects. Various approaches~\cite{wang2022fourier, shao2023tensor4d, cao2023hexplane, fridovich2023k} extend NeRF to dynamic scenes, but they fall short of supporting long sequences and streaming. Some~\cite{wang2023neural, song2023nerfplayer, li2022streaming} can stream radiance fields for each frame, but the heavy computational overheads make them impractical on mobile devices. 
Notably, the recent VideoRF~\cite{wang2024videorf} can stream and render dynamic radiance fields on mobiles, but the rendering results still suffer from blurriness.

Only recently, the 3D Gaussian Splatting (3DGS)~\cite{kerbl3Dgaussians} achieves exceptional rendering ability at speed and quality. While the original 3DGS only supports static scenes, we have witnessed its rapid adoption into dynamic scenes. Many works~\cite{li2024spacetime, wu20244d, sun20243dgstream} extend 3DGS to dynamic scenes with high-quality playback, but these methods do not support long sequences due to the per-frame storage overhead. The animatable approaches~\cite{Zielonka2023Drivable3D,zheng2023gpsgaussian,hu2024gaussianavatar} can handle extended human motions but are still too computationally intensive to be employed on mobile devices. On the other hand, thanks to the explicit representation, researchers explore implementing the Gaussian rasterization and rendering on various light-weight web platforms~\cite{antimatter15_splat,42yeah_rasterizing_splats}, with the aid of effective attribute compression of static scenes~\cite{fan2023lightgaussian,niedermayr2023compressed,lee2024compact}. Yet, their computing and storage overheads are still insufficient for streaming dynamic Gaussians on the fly, especially for real-time interactions.
In a nutshell, it remains unsolved to stream and render dynamic 3DGS for viewing volumetric videos on mobile devices. 
It requires an efficient and effective training process to generate compact Gaussian representations. Also, the generated Gaussian assets must be compatible with hardware video codecs and shader rendering on mobile devices for efficient rendering.

To tackle the above challenges, in this paper, we propose V\textsuperscript{3}, a novel approach for streaming and rendering dynamic 3DGS on mobile devices. As shown in Fig.~\ref{fig:teaser}, it enables viewing high-quality volumetric videos with real-time and immersive interactions. 
Our key idea is to formulate the dynamic Gaussian sequence as a compact 2D Gaussian video, which naturally supports hardware video codecs for efficient streaming and decoding. Temporally, the Gaussian attributes in each frame are mapped to multiple dimensions of a 2D video pixel. During rendering, we can efficiently extract 3D Gaussian attributes (e.g., rotation, scale, position, opacity, color, and spherical harmonics) from each pixel in the 2D plane. 

Furthermore, we present an efficient training scheme to acquire our 2D dynamic Gaussian representation from multi-view video footage. It can maintain the temporal consistency of the 2D Gridded Gaussians to reduce the storage requirements. Specifically, we separate the entire sequence into frame groups to handle topological changes and long-duration sequences. Within each group, we optimize the first frame with static 3DGS and prune to reduce storage, then conduct sequential training in a two-stage manner frame by frame, to separately and accurately model the motion and appearance attributes. In the first stage, given the optimized Gaussians from the previous frame, we adopt hash encoding and a shallow MLP to efficiently estimate the relative position changes of each Gaussian splat. Secondly, we fine tune the other Gaussian attributes with a residual entropy loss and a temporal loss to enhance the inter-frame consistency. For the former, we adapt the entropy loss technique from previous static methods~\cite{hac2024,zhang2024efficient} into our inter-frame residuals of Gaussian attributes. It reduces the entropy of Gaussian attributes to improve robustness to quantization for compression. The latter temporal loss further enhances the temporal consistency of our 2D Gridded Gaussians.
We adopt Morton sorting to pack all the Gaussian attributes into a 2D format. It ensures that neighboring Gaussian points in 3D space remain neighbors in the 2D representation, thereby enhancing the spatial consistency for subsequent video codec processing.

When our player receives the 2D Gaussian stream, it can leverage hardware video codecs for rapid decoding and utilize shaders for real-time rendering.
In this way, we can achieve high-quality volumetric video streaming even on mobile devices. 
As shown in Fig. ~\ref{fig:teaser}, users enjoy browsing volumetric videos of interest anytime, anywhere, just like on YouTube. Thanks to our compact representation, users can instantly share their favorite volumetric videos with friends, conveying their joy and immersion instantly.

In summary, our primary contributions are as follows:
\begin{itemize}
\item We propose V\textsuperscript{3}, a novel approach to support rendering volumetric video on common devices via streaming Gaussian splats with high quality.
\item We present a compact dynamic Gaussian representation that bakes Gaussian attributes into a 2D Gaussian Video to facilitate hardware video codecs.
\item We propose an efficient training strategy that maintains temporal continuity through motion-appearance disentanglement, residual entropy loss, and temporal loss.
\item We propose multi-platform volumetric video players, supporting real-time playing and streaming. 
\end{itemize}

\section{Related Work}

\paragraph{Novel View Synthesis for Dynamic Scenes.}

Recent methods modeling dynamic scenes including ~\cite{pumarola2021d, tretschk2021non, li2021neural, park2021nerfies, song2023nerfplayer} that predict motion with a deformation field, ~\cite{gao2021dynamic, li2022neural, zhang2021editable, jiang2023instant} use time-aware neural network, ~\cite{fang2022fast, wang2022fourier, wang2023neus2} utilize explicit voxel representation, ~\cite{cao2023hexplane, shao2023tensor4d, fridovich2023k} use multi-plane decomposition, ~\cite{zhao2022human} use animated mesh for efficient rendering, ~\cite{peng2021neural} that use SMPL driven latent code, ~\cite{wang2021ibutter} that based on IBR(Image Based Rendering). Although these approaches have made some progress in modeling dynamic scenes, they are still limited by their rendering quality as well as rendering and optimization speeds.

Recently, with explicit representations achieving high-quality rendering and fast training in static scenes, several methods~\cite{luiten2023dynamic, yang2024deformable, wu20244d, jiang2024hifi4g, li2023animatablegaussians, li2024spacetime, zheng2023gpsgaussian} utilize Gaussian splatting~\cite{kerbl3Dgaussians} to model dynamic scenes, achieving high rendering quality. However, these methods are constrained by their large model sizes and long training durations, making it challenging to quickly generate and efficiently transfer volumetric video across platforms. By learning motion between adjacent using Hash MLP, our method achieves fast optimization speed for dynamic sequences, enabling fast generation of dynamic assets. 

\paragraph{Efficient Radiance Field.} 
Neural Radiance Fields (NeRF)~\cite{mildenhall2021nerf} represent scenes as implicit neural networks, capturing high-quality scene details but requiring several hours to train and lacking support for real-time rendering. To accelerate training and rendering, methods including~\cite{liu2020neural, muller2022instant, yu2021plenoctrees, sun2022direct} convert implicit structures to explicit voxel, tri-plane or mesh representations combined with tiny MLPs, significantly speeding up scene training and rendering. Some methods, including~\cite{li2024nerfcodec,chen2021nerv,li2023compressing} use serval methods to compress the neural radiance fields, making it memory-efficient. However, implicit representations in these methods limit faster rendering speeds. Recently, 3D Gaussian Splatting~\cite{kerbl3Dgaussians} has achieved rendering speeds of several hundred FPS by using fully explicit Gaussian point clouds to represent scenes, along with fast optimization capabilities. To address the large storage demands of explicit point cloud representations, some methods~\cite{niedermayr2023compressed, fan2023lightgaussian} employ techniques like codebooks, point reduction, and entropy encoding to compress models by 20 times. However, these methods face challenges in fast model decoding. Compact-SOG~\cite{morgenstern2023compact} restore Gaussians in 2D images and compress using image compression technique, but cannot exploit the temporal continuity. In contrast, our approach retains explicit representations and encodes dynamic Gaussians using hardware codec, achieving lower storage and real-time decoding and streaming. 


\paragraph{Cross Device Neural Radiance Field Rendering.}
Many methods have attempted to render radiance fields on lightweight platforms such as mobile devices. Specifically ~\cite{chen2023mobilenerf, yariv2023bakedsdf, tang2023delicate} represent scenes using explicit triangle meshes combined with neural textures, making them compatible with traditional rendering pipelines. ~\cite{cao2023real} reduces rendering computation by distilling NeRF into a light field, thus improving rendering efficiency. ~\cite{reiser2023merf}, based on voxel representation, achieves efficient rendering on mobile platforms by decomposing scenes into three planes.
Recently, 3D Gaussian Splatting~\cite{kerbl3Dgaussians} has used explicit point clouds to represent scenes and has combined this with fast rasterization rendering techniques that are compatible with traditional rendering pipelines, which further accelerates rendering, enabling high FPS rendering on mobile platforms. 

\paragraph{Streamable Volumetric Video}
Recent methods have proposed volumetric video streaming, aimed at enabling real-time viewing of high-quality volumetric videos across multiple platforms. Among these, StreamRF~\cite{li2022streaming} employs a time-dependent sliding window to model dynamic scenes, but its representation is not conducive to efficient streaming. ReRF~\cite{wang2023neural} introduces an FVV codec specifically designed for streaming, achieving efficient compression and transmission of dynamic scenes. VideoRF~\cite{wang2024videorf} and TeTriRF~\cite{wu2024tetrirf} leverage video codecs to exploit temporal redundancy in dynamic scenes, where VideoRF maps 3D voxel grid data to 2D images via a mapping table, while TeTriRF decomposes the scene’s appearance features into tri-planes and stores these planes as 2D images. However, these methods are constrained by the rendering efficiency of implicit representations, making them challenging to deploy on mobile platforms.

With the introduction of 3DGS~\cite{kerbl3Dgaussians}, recent approache 3DGStream~\cite{sun20243dgstream} represent scenes using keyframes, MLP-based motion transformations, and newly added 3D Gaussian point clouds, enabling fast rendering and training. Nevertheless, during playback, the need to store and query the MLP hinders efficient streaming and mobile platform rendering.
By encoding the explicit 3DGS point clouds using video codecs, our methods facilitate efficient transmission and enable fast rendering even on mobile platforms.

\begin{figure}[t]
    \centering
    \includegraphics[width=\linewidth]{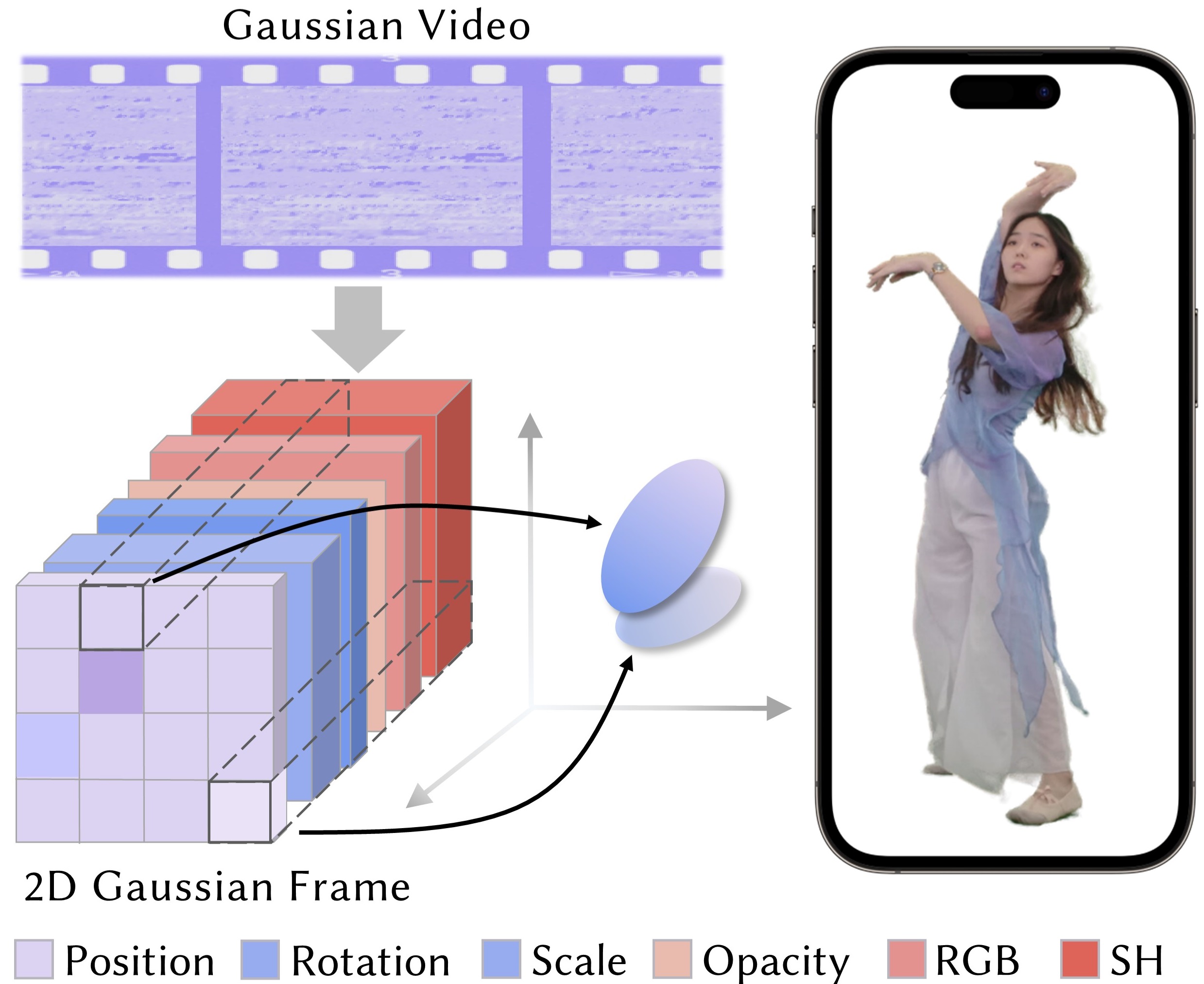}
    \vspace{-5mm}
    \caption{We model dynamic 3DGS as a 2D video with multiple dimensions, where each frame corresponds to its specific 3DGS attributes. During the rendering, we extract Gaussian properties from each pixel to recover Gaussian Splat structural.}
    \vspace{-3mm}
    \label{fig:rep}
\end{figure}

\section{V\textsuperscript{3} Representation}


Given a set of multiview video sequences of human performance, we aim to ensure that everyone can seamlessly experience high-quality free-viewpoint video (FVV) rendering on any device, anywhere. 
To achieve this, we model the dynamic 3DGS sequence as a compact 2D Gaussian video, which naturally supports hardware video codecs for efficient streaming and decoding. 
By utilizing shader-based rendering, we can achieve efficient rendering on various portable devices.

As illustrated in Fig. ~\ref{fig:rep}, we treat the dynamic 3DGS~\cite{kerbl3Dgaussians} as multiple 2D videos. Define $R_t, S_t,x_t,o_t,c_t, SH_t$ as the rotation, scale, position, opacity, color, and spherical harmonics of the Gaussian splats at time $t$, then the total number of dimensions of 3DGS is $N$ and 
\begin{equation}
    N = d(R_t) +  d(S_t) + d(x_t) + d(o_t) + d(c_t) + d(SH_t), 
\end{equation}
where function $d$ is the number of specific attributes. 
The 2D videos can be denoted as $\{V^i\}_{i = 0}^N$, each 2D video corresponds to a single dimension of attributes of 3DGS, so each frame $\mathbf{I}^i_t$ corresponds to the attributes of the 3DGS scene at frame t, and finally the pixel value at the same position of each frame corresponds to the splat's attribute values. During rendering, given the current frame index t, we can construct the current 3DGS scene from videos. Firstly, we extract 2D encoded images $\{\mathbf{I}^i_t\}_{i = 0}^N$ from videos, then we synchronously traverse the pixel coordinates u,v across all images to construct each splat, as shown in the following equation:
\begin{equation}
        x_t,R_t,S_t,o_t,c_t,SH_t = \varphi(\{\mathbf{I}^i_t[u, v]\}_{i = 0}^N),
\end{equation}
where $\mathbf{I}_t^i$ are the 2D encoded images at time $t$, $[u, v]$ is the 2D index in pixel coordinate,  $R_t, S_t,x_t,o_t,c_t, SH_t$ is the attributes of the Gaussian splats at time $t$. Here the mapping function $\varphi$ refers to the synchronous traversal for all pixels across all images. 
Due to the unordered nature of point clouds, there is no need to construct 3DGS in a specific order; instead, we can build the 3DGS directly according to the order in which pixels are read, which eliminates the need to record a mapping table, significantly enhancing the decoding speed.

\begin{figure}[t]
    \centering
    \includegraphics[width=\linewidth]{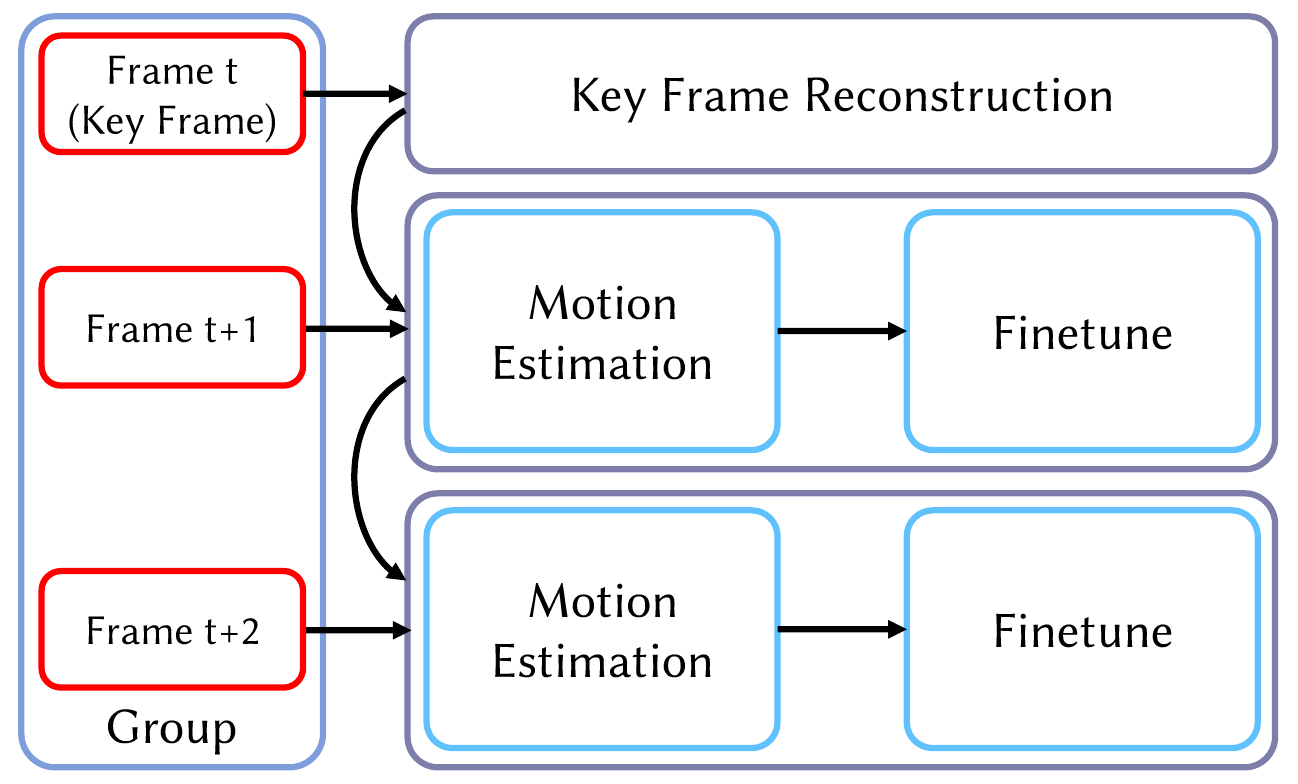}
    \caption{\textbf{Overview of V\textsuperscript{3} training.} For a frame group, we select the first frame as the keyframe and reconstruct it with a prune fine tune strategy to control the number of Gaussians. For other frames in the frame group, we employ the sequential two-stage training strategy for each frame to get the per-frame 3DGS model. 
    }
    \vspace {-3mm}
    \label{fig:algorithm_overview}
\end{figure}

As 3DGS ~\cite{kerbl3Dgaussians}, each Gaussian splat is projected through
\begin{equation}
\Sigma=R S S^T R^T, \quad
\quad \Sigma^{\prime}=J W \Sigma W^T J^T.
\end{equation}


Here, $\Sigma$, $\Sigma^{\prime}$ is the covariance matrix in world and camera coordinate. $R$ is the rotation matrix and $S$ is the scale matrix. $J$ is the Jacobian of the affine approximation of the projective transformation, and $W$ is the viewing transformation matrix. With the covariance matrix, we can get the projected Gaussians for rendering, and the color of a pixel on the image plane can be calculated by alpha blending the N-ordered Gaussians from close to far: 

\begin{equation}
C=\sum_{i \in N} c_i \alpha_i^{\prime} \prod_{j=1}^{i-1}\left(1-\alpha_j^{\prime}\right),
\end{equation}

where the computed Gaussians are overlapping the pixel, $c_i$ is the view-dependent color of each 3D Gaussian, $\alpha_i$ is the opacity multiplied by the 3D Gaussian and the corresponding covariance matrix $\Sigma^{\prime}$. By leveraging our 2D Gaussian video representation, we can provide fluent Gaussian splats streaming with high fidelity on common devices.

\begin{figure}[t]
    \centering
    \includegraphics[width=\linewidth]{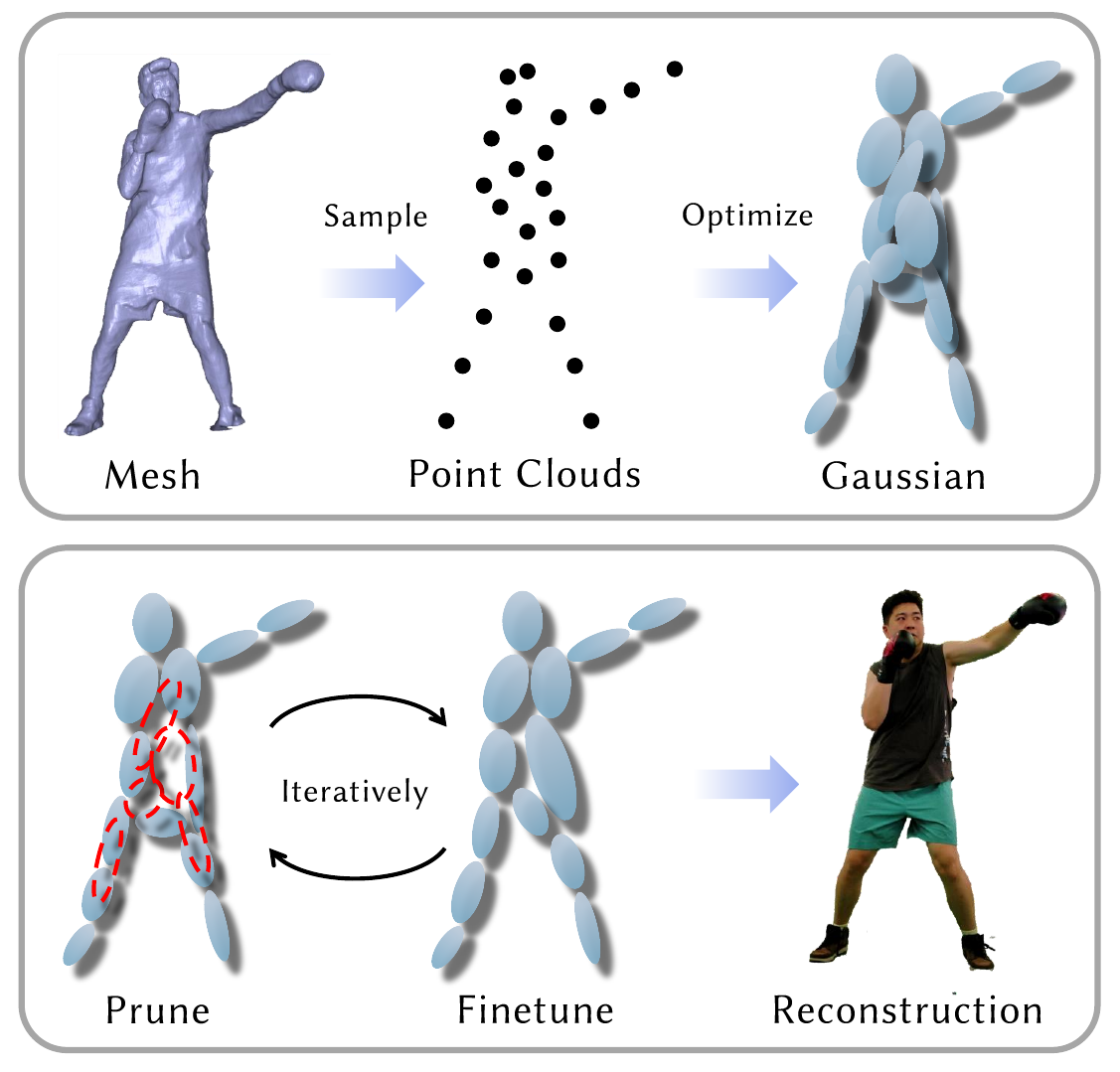}
    \caption{\textbf{Keyframe training.} Our keyframe uses the triangle mesh generated by NeuS2~\cite{wang2023neus2} as the initial point cloud and then constructs the Gaussian Splatting model. To make our representation more compact, we further prune the Gaussians according to opacity and fine tune. By iterative pruning and fine tuning, we can efficiently control the storage of our model. 
    }
    \vspace {-3mm}
    \label{fig:keyframe}
\end{figure}

\begin{figure*}[t]
    \centering
    \includegraphics[width=\textwidth]{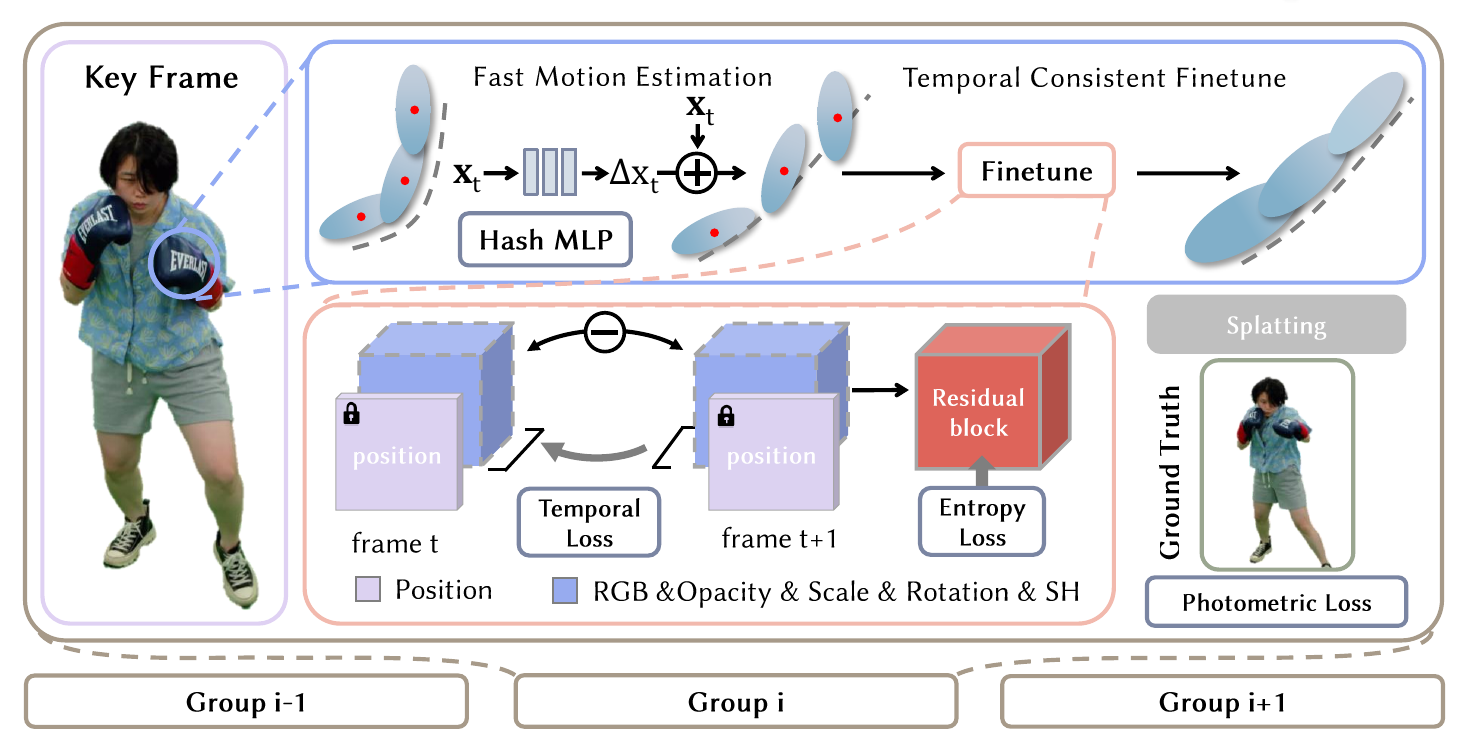}
    \caption{\textbf{Two-stage training.} First, we divide the long sequences into groups for training. In the first stage, we use hash encoding following a shallow MLP with position as input to estimate the motion of the human subjects. In the second stage, we fine tune the attributes of the warped Gaussians from stage 1 with residual entropy loss and temporal loss, which yields 2D Gaussian video with high temporal consistency and thus we can use a video codec to perform efficient compression. 
    }
    \vspace {-3mm}
    \label{fig:pipeline}
\end{figure*}

\section{V\textsuperscript{3} Reconstruction}
Acquiring our compact V\textsuperscript{3} representation for fluent streaming with high quality is challenging. 
Using a per-frame training method trivially results in significant storage requirements, even after video codec compression. 
This is because it fails to account for the continuity of Gaussian properties between adjacent frames. 
Additionally, many Gaussian attributes from the previous frame can be reused, eliminating the need for retraining from scratch.
To address this, as shown in Fig. ~\ref{fig:algorithm_overview}, we propose an efficient grouped training scheme to generate our compact V\textsuperscript{3} representation while maintaining temporal consistency.
Furthermore, we propose a temporal entropy loss and a temporal loss to further enhance the temporal continuous on our V\textsuperscript{3} representation. 

\subsection{Grouped V\textsuperscript{3} Training}
Given a sequence of dynamic human-centric performances, we separate them into frame groups to support dynamic scenes with topological transformations and infinite lengths. For the balance of training speed of speed and quality, we set the frame group size to 20. In a frame group, we set the first frame as the keyframe, then perform static 3DGS reconstruction and prune the point cloud to reduce storage. For the rest frames, we estimate the motion from the last frame and then fine tune the warped model. By sequentially optimization for each frame, we could efficiently obtain the temporal consistent 3DGS model of each frame. 

\paragraph{Key Frame Reconstruction}
As shown in Fig. ~\ref{fig:keyframe}, We chose the first frame as the key frame for each frame group and sample from mesh generated by NeuS2~\cite{wang2023neus2} to obtain the initial point cloud, then we optimize Gaussian Splatting for the keyframe. To reduce storage, we control the number of splats to be under 100k by removing points with low opacity. We sort splats of the keyframe according to the opacity, pruning the points, and fine tuning the attributes without densifying or cloning. According to compact-SOG~\cite{morgenstern2023compact}, pruning 30\% Gaussians with the lowest opacity will not affect the quality of scenes, so we set the prune ratio to 30\%. By iteratively pruning and fine tuning, we can control the number of points while maintaining rendering quality. 

Next, as shown in Fig.~\ref{fig:pipeline} we apply a sequential training scheme to generate the Gaussian representation of the subsequent frame through motion estimation and fine tuning stages. 

\paragraph{Fast Motion Estimation.}
Within each frame group, we strive to maintain the temporal continuity of the final baked 2D Gaussian Video to achieve compact storage. 
To improve this consistency, we aim to map the Gaussian primitives of the same surface to the same encoded pixel positions across different frames as much as possible. To achieve this, we use hash encoding with a shallow MLP to quickly model the position changes of Gaussian primitives sequentially over time.

By doing so, we maintain a constant number of Gaussian primitives across different frames. 
These temporally varying Gaussians can be mapped to the same encoded pixel position using the same mapping operation within the frame group. 
While Gaussian densification and pruning can model the appearance changes of dynamic scenes, they lose the motion estimation of the Gaussian primitives, and the varying number of Gaussians between frames can significantly disrupt the continuity after mapping, potentially introducing jitter artifacts.

We use a multiresolution hash grid with a shallow MLP to estimate the position change of each Gaussian primitive. Given a hash grid with resolution level $L$
, we can get its hash encoding as 
\begin{equation}
    h(x) = \{h^{1}(x), h^{2}(x), ..., h^{L}(x)\}
\end{equation}
Then with a shallow neural network denoted as $MLP$, we can efficiently estimate the motion by 
\begin{equation}
    \Delta x=MLP(h(x_{t-1})),
\end{equation}
given the position $x$ of frame t-1 with its multiresolution $h(x_{t-1})$. Finally, we can achieve the position of frame t by $x_t=x_{t-1}+ \Delta x$.

Following 3DGS, the loss function of this stage for each frame in order is:
\begin{equation}
\mathcal{L}=(1-\lambda) \mathcal{L}_{\text{photometric}}+\lambda \mathcal{L}_{\text {D-SSIM }},
\end{equation}
where $\mathcal{\lambda}=0.2$.

In this way, we can estimate the position changes between two frames within seconds, significantly reducing training time and minimizing redundant retraining. 
Simultaneously, we can generate the correspondence between Gaussian splats across frames, reducing the attribute differences between frames and enabling better compression.

\subsection{Temporal Regularization}


Since we will use video codecs to compress our V\textsuperscript{3} representation, the residuals between frames are stored in bitstreams after entropy encoding. To address this, during the fine tune stage, we propose a temporal loss to enhance temporal continuity, thereby reducing the residuals of Gaussian attributes between frames. 
Additionally, inspired by ~\cite{hac2024,zhang2024efficient}, we apply the entropy loss on the residuals to exhibit low entropy and make the residuals between frames robust to quantization.
\paragraph{Residual Entropy Loss}

In the entropy encoding process, we encode data according to its probability. If a value appears more frequently, we use less bits to encode it.
Therefore, the entropy loss aims to enforce data to be closer to the center of its distribution, increasing the probability of repeated values, which leads to less storage.
 From an information theory perspective, this approach minimizes entropy, enhancing compressibility.
 
\begin{figure}
\centering
\includegraphics[width=\linewidth]{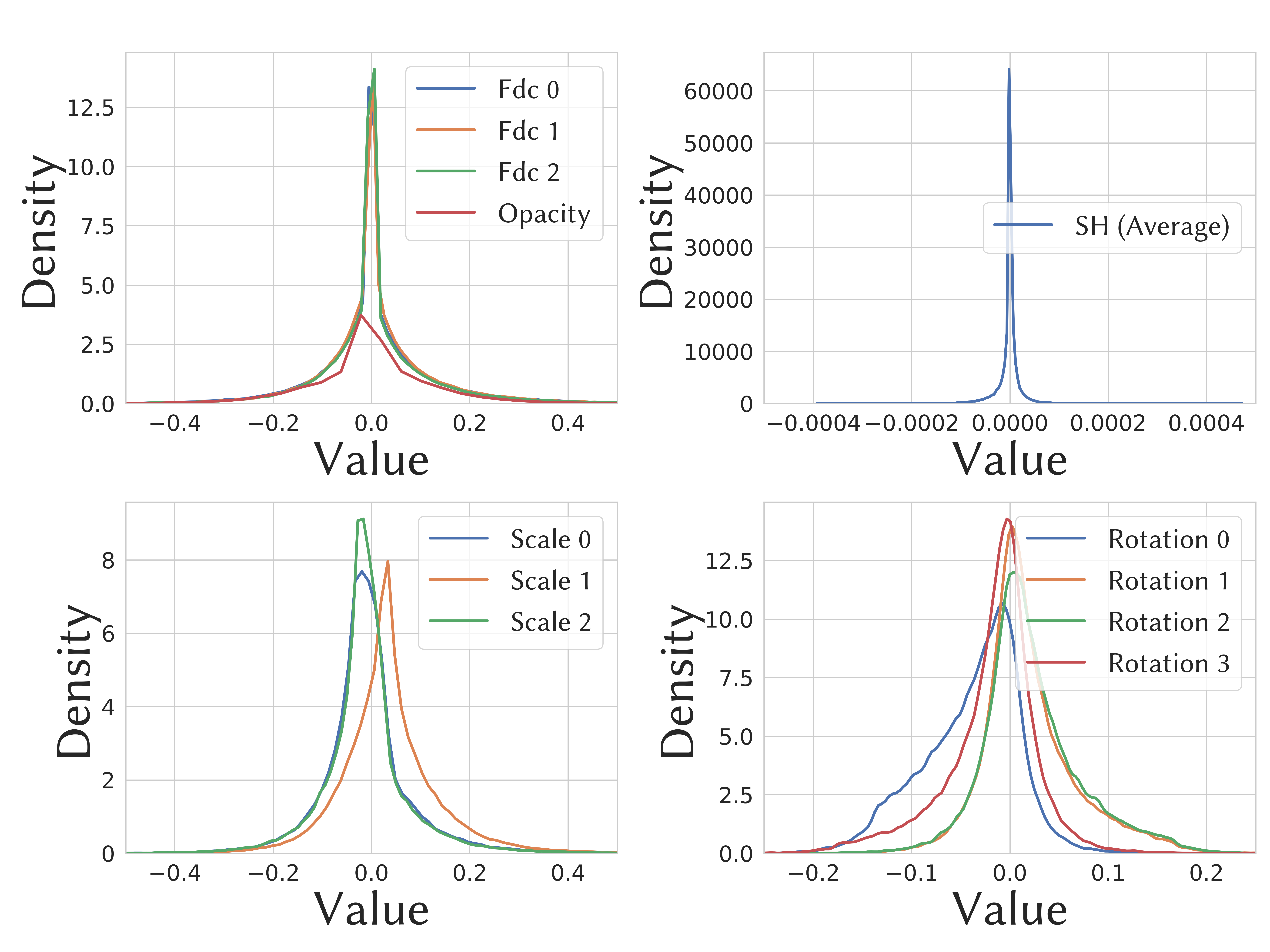}
\caption{Analysis of the residual Gaussian attribute distribution revealed that the residuals in appearance, scale, and rotation exhibit Gaussian characteristics.}
\vspace {-3mm}
\label{fig:entropy}
\end{figure}

To reduce entropy, we use a loss function to constrain the data distribution. 
According to statistics, as shown in Fig. ~\ref{fig:entropy}, we find that the residuals of rotation, scale, opacity, and spherical harmonics between frames are approximately Gaussian distributed. Therefore, we assume these attributes follow Gaussian distributions with their individual learnable $\mu$ and $\sigma$. 

Specifically, let $y$ represent one of the Gaussian properties, $y_t \in [R_t,S_t,o_t,c_t, SH_t]$, $q_i$ is the quantization region of different Gaussian attributes, we can calculate the quantized residual as:
\begin{equation}
    \Delta y_t= y_t-y_{t-1},
\end{equation}

\begin{equation}
    \hat{\Delta y_t}= (\Delta y_t-y_t^{min})/(y_t^{max}-y_t^{min})*q_i
    +\mathcal{U}\left(-\frac{1}{2}, \frac{1}{2}\right).
\end{equation}
where $\mathcal{U}\left(-\frac{1}{2}, \frac{1}{2}\right)$ is a random uniform noise on the residuals to simulate rounding operation, thereby increasing the robustness of our parameters to quantization.

Then, we can approximate the probability mass function (PMF) of our residuals by calculating the difference between two cumulative distribution functions (CDFs) as follows:
\begin{equation}
    P(\hat{\Delta y_t})= P_{c d f}\left(\hat{\Delta y_t}+\frac{1}{2}\right)-P_{c d f}\left(\hat{\Delta y_t}-\frac{1}{2}\right).
\end{equation}

Consequently, our entropy loss is calculated as the summation of bit consumption as:
\begin{equation}
    \mathcal{L}_{\text {entropy}}=\frac{1}{N} \sum_{y_t \in \{R_t,S_t,o_t, c_t,SH_t\}}-\log _2\left[P\left(\hat{\Delta y_t}\right)\right],
\end{equation}
where $N$ is the number of Gaussian splats.

In this way, the residual values of each Gaussian attribute are closer to the center of their Normal distributions, thereby reducing temporal entropy and making our V\textsuperscript{3} robust to quantization, maintaining high quality even at low bitrates.

\paragraph{Temporal Loss.}
To further improve the temporal coherence of our V\textsuperscript{3}, we apply a temporal loss to minimize the difference between adjacent frames in the fine tuning stage.
During the sequential training,  we improve inter-frame similarities by using the attributes of the previous Gaussian image to regularize the current Gaussian image as

\begin{equation}
\mathcal{L}_{\text {temp}}=\frac{1}{W\times H} \sum_{y_i \in \{R_i,S_i,o_i,c_i, SH_i\}}\left\|y_t-y_{t-1}\right\|_1,
 \end{equation}
where $W$ and $H$ are the width and height of our Gaussian Video, and $p$ is the pixel value at the dimensions of rotation, scale, opacity, color, and SH. 
By applying temporal smoothness in this way, we can further reduce residuals during the video codec process, thereby conserving storage.

\paragraph{Total Loss.} In our fine tuning stage, our total loss function for each frame in order is formulated as follows:
\begin{equation}
\mathcal{L}=(1-\lambda) \mathcal{L}_{\text{photometric}}+\lambda \mathcal{L}_{\text {D-SSIM }}+\lambda_e \mathcal{L}_{\text {entropy}} + \lambda_t \mathcal{L}_{\text {temp}},
\end{equation}
where $\lambda_e$ and $\lambda_t$ are the weights for our regular terms. 
\subsection{Baking 3D Gaussians into a 2D Format.}
To utilize video codec for compression and streaming,  we need to bake the Gaussians into a 2D format for compression and streaming. Benefiting from the unordered and unstructured nature of point clouds, unlike the voxel grid representation used in VideoRF~\cite{wang2024videorf}, we do not need to store a 2D to 3D mapping table to reconstruct the 3D spatial structure. Instead, we only need to ensure that different attributes of 
the same Gaussian are stored at the same index position in different images, which allows us to easily reconstruct the point cloud structure without requiring additional storage space for correspondence. 

\paragraph{Quantization bits setting}
Since Gaussians are sensitive to the position of the point cloud, we require more precise position information. Thus, we quantize the position attribute using uint16, while other attributes are quantized using uint8. Since we use uint8 PNG images to store attributes, we split the position information into two uint8 values, representing the high and low 8 bits of uint16, respectively. We ensure lossless compression for the high uint8 bits to maintain precision. 

\paragraph{2D Encoded Image Resolution}
To better store the Gaussians, we adopt an adaptive resolution 2D format. The number of Gaussians is consistent across frames within the same frame group. When performing video compression, storing the Gaussians in a 2D format with an edge length of 8 is more codec-friendly, further reducing storage requirements. We determine the smallest square 2D format that can accommodate all the Gaussians while ensuring the square’s edge length is a multiple of 8, thereby minimizing empty grid space. This approach ultimately results in a codec-friendly video compression format.

\paragraph{Compression QP Setting}
In addition, in video codec, the impact of compression on the rendering quality of Gaussians varies across different attributes. In the H.264 codec, we can adjust the QP (Qstep) coefficient to control the compression rate. A higher QP results in greater compression loss and reduced storage requirements. For each attribute, we control the other attributes while gradually increasing the QP for the target attribute to observe whether the rendering quality degrades sharply once the QP exceeds a certain threshold. Experimental results indicate that when the QP exceeds 22, significant compression losses occur in the Gaussian RGB, Scale, and Rotation attributes. Therefore, we introduce a threshold-based adjustment strategy. When the QP is less than 22, all attributes are compressed using the same QP value. When the QP exceeds 22, the RGB, Scale, and Rotation attributes are compressed with a fixed QP of 22, while the QP for the remaining attributes can be further increased. This approach allows us to achieve additional compression and reduce storage while maintaining acceptable rendering quality.

\paragraph{Morton Sort}
Inspired by SOG~\cite{morgenstern2023compact} 
, we use Morton Order to sort the positions of the point cloud and then bake the Gaussians into a 2D format. In detail, for an unordered 3D Gaussian point cloud, we convert the coordinates of each point from float to integers with a range of $2^{21}$, then calculate their Morton codes and sort them. Since Morton sorting maps spatially close points to adjacent positions on 2D encoded images, and considering points close to each other in a 3D Gaussian have similar properties, Morton sorting effectively places similar points closely on the image, enhancing the spatial consistency of the 2D encoded image, which is more conducive to video codec processing.

\begin{figure}
\centering
\includegraphics[width=\linewidth]{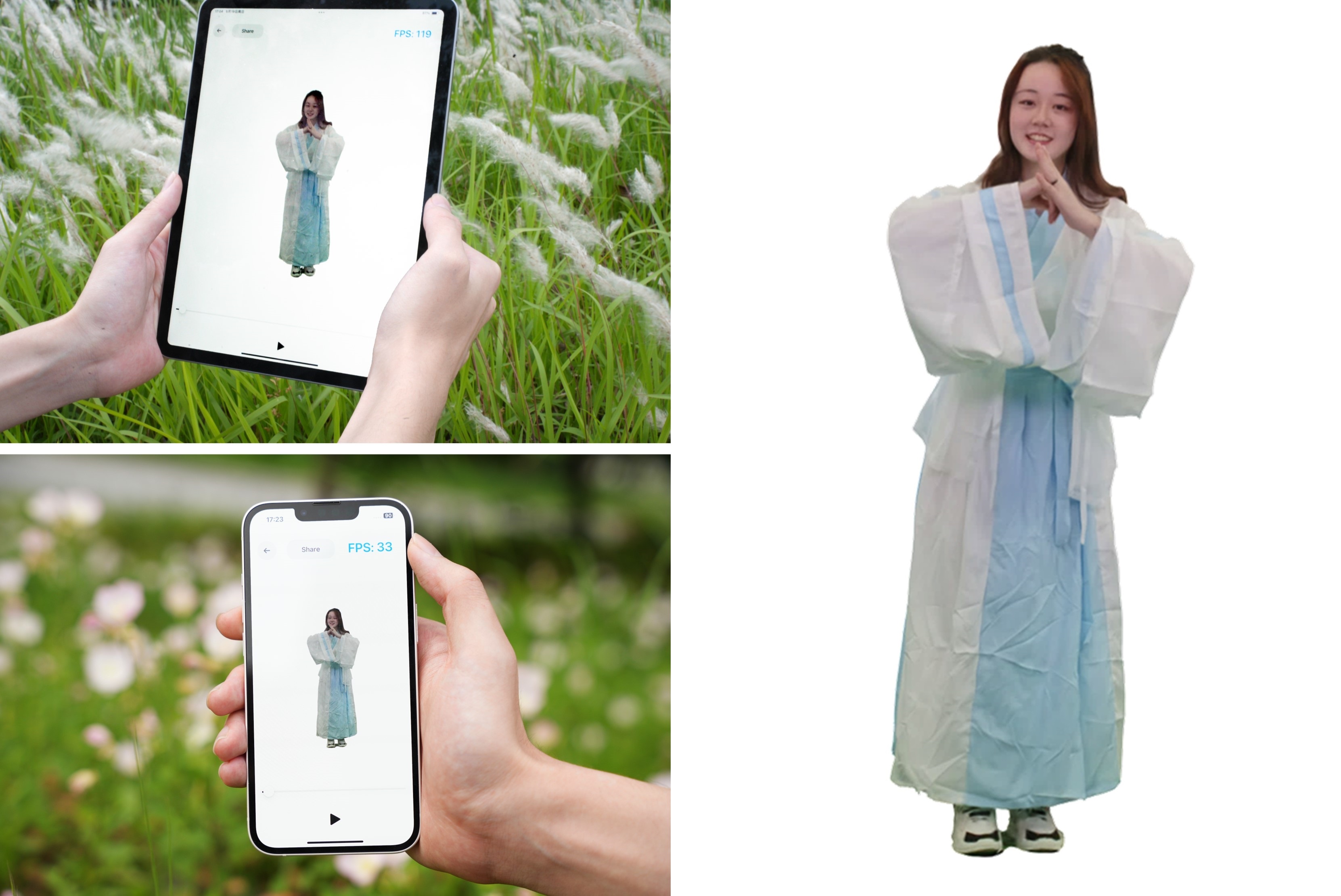}
\caption{We support high-quality rendering on various mobile platforms anytime and anywhere, enabling real-time streaming and rendering in diverse environments.}
\vspace {-5mm}
\label{fig:application}
\end{figure}

\begin{figure*}[p]
    \centering
    \includegraphics[width=\linewidth]{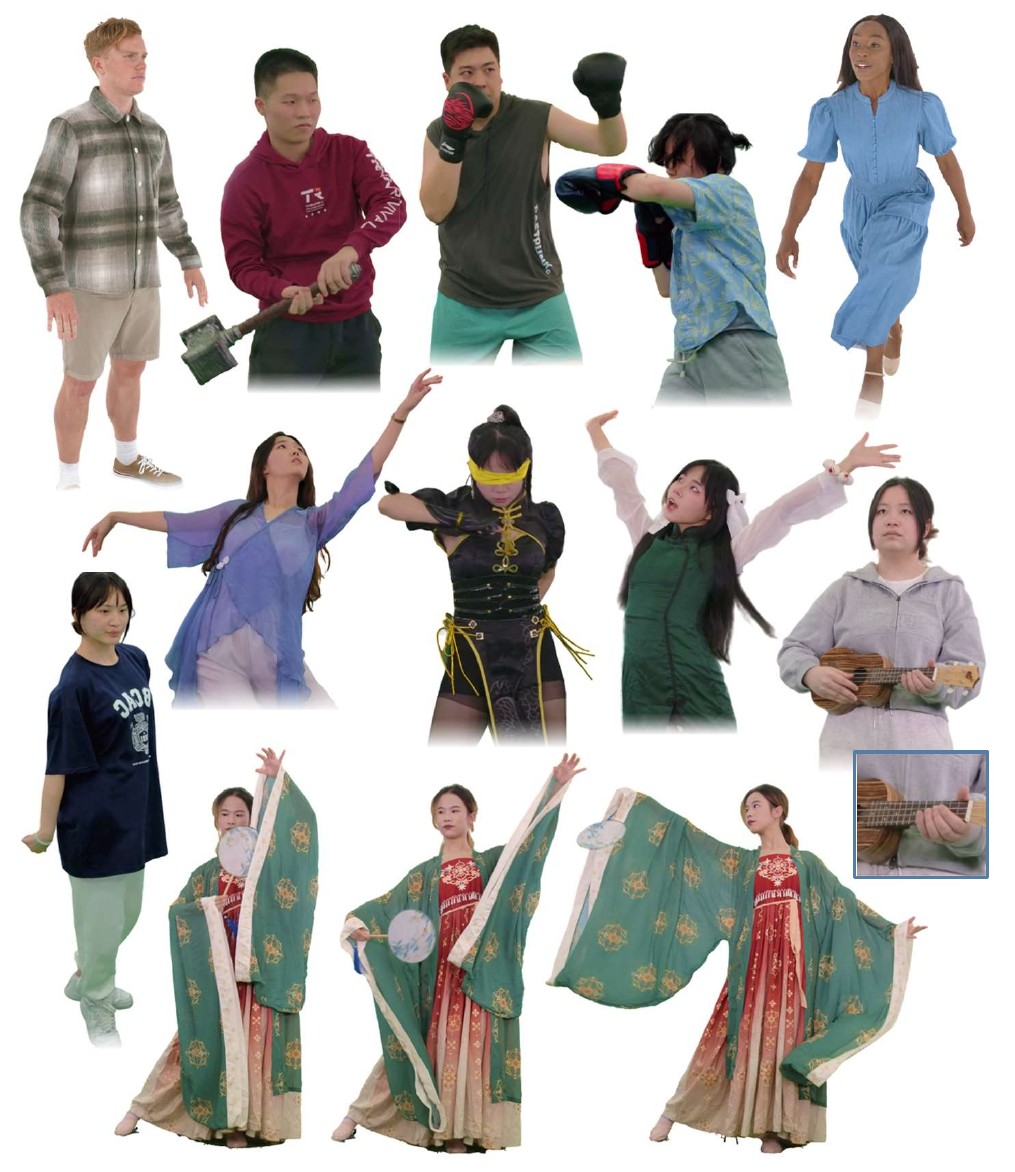}
    \caption{Gallery of our results. Our method can achieve high-quality novel view synthesis in scenes with challenging motion and flexible topology changes. }
    \label{fig:gallery}
\end{figure*}

\begin{figure*}[t]
    \centering
    \includegraphics[width=\linewidth]{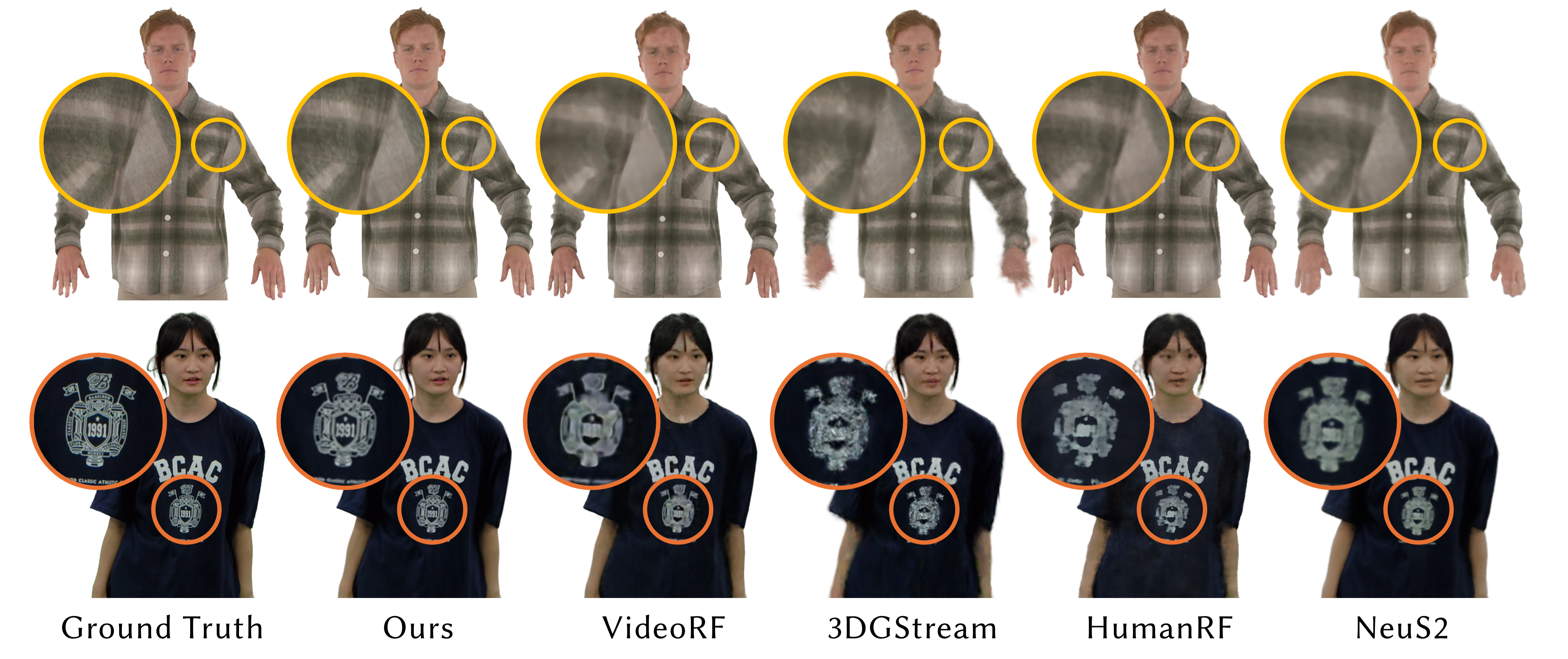}
    \caption{Qualitative Comparison against recent SOTA methods including VideoRF~\cite{wang2024videorf}, 3DGStream~\cite{sun20243dgstream}, HumanRF~\cite{icsik2023humanrf}, NeuS2~\cite{wang2023neus2}. Our method achieves high-quality rendering with clear details. }
    \label{fig:comparison}
\end{figure*}

\section{V\textsuperscript{3} Player}

We implement our V\textsuperscript{3} player on various platforms, including desktop, laptop, and mobile devices. For the baked 2D Gaussians, we use FFmpeg to utilize H.264 codec to obtain several 2D Gaussian videos and upload them to the resources server. For the decoding process, we fetch the Gaussian video streams and use OpenCV for hardware decoding of H.264 video to obtain 2D Gaussian images. On Desktop and Laptop platforms, we utilize multithreading: thread 1 handles video fetching and decoding, while thread 2 performs inverse quantization to the decoded images and reconstructs them into Gaussian point cloud structures for rendering. On Mobile devices, we implemented the viewer using Swift, also employing a multithreading strategy. To maximize hardware resource utilization, we use Metal's compute shaders to transform the images back into Gaussian point clouds. For the rendering pipeline, we use Metal Shaders to implement alpha blending, thereby eliminating the need for CUDA devices and enabling rendering on cross-device.
Our V\textsuperscript{3} player supports streaming volumetric video from the network and real-time decoding and rendering, including features like free view angle adjustment, timeline scrubbing, and play/pause operations. As shown in Fig. ~\ref{fig:application}, we support free viewing of volumetric videos on diverse mobile devices, enabling an immersive, high-quality volumetric video viewing experience anytime and anywhere.


\section{Experimental Results}
We evaluated the reconstruction performance of V\textsuperscript{3} across multiple scenes, on the dataset of ReRF, Actors-HQ, and newly captured dynamic data at 30 fps and 3840 × 2160 resolution with 81 views. We trained the model using a single NVIDIA GeForce RTX3090. In stage 1, we set 16 levels and 4 features per level for hash grid, 64 neurons per layer, and 2 hidden layers for shallow MLP. In stage 2, we set the weights of the regular terms with $\lambda_e$ as 1e-4 and $\lambda_t$ as 1e-3 for all sequences. As shown in Fig. ~\ref{fig:gallery}, V\textsuperscript{3} achieves superior reconstruction results in various complex human scenes. It can handle intricate movements such as dancing, boxing, and animated character actions. We support real-time streaming to render long sequences of FVV videos on human-centric scenes, providing an immersive viewing experience of human dynamic performances. Please refer to the supplementary video for more video results.
\begin{table}[t]
\begin{center}
\footnotesize
\centering\setlength{\tabcolsep}{6pt}
\renewcommand{\arraystretch}{1.1}
\setlength{\tabcolsep}{1.5mm}{\begin{tabular}{l | c | cccc }
\multicolumn{6}{c}{ \colorbox{red!25}{\strut best} \colorbox{orange!25}{\strut second-best} } \\
\hline
 Dataset & Method & PSNR$\uparrow$ & SSIM$\uparrow$ & Training Time(Min) $\downarrow$ & Size(MB)$\downarrow$ \\ 
\hline
\multirow{5}{*}{ReRF}
 & HumanRF   & 28.82 & 0.900 & 1.83 & 2.800 \\
 & NeuS2     & 28.48 & \SecondBestCellColor{0.977} & 1.62 & 29.07 \\ 
 & VideoRF   & \SecondBestCellColor{32.01} & 0.976 & >20  & \SecondBestCellColor{0.658} \\
 & 3DGStream & 27.26 & 0.960 & \BestCellColor{0.12} & 7.644 \\ \cline{2-6}
 & Ours      & \BestCellColor{32.97} & \BestCellColor{0.983} & \SecondBestCellColor{0.82} & \BestCellColor{0.532} \\ 
\hline
\multirow{5}{*}{Actors-HQ}
 & HumanRF      & 30.14 & \BestCellColor{0.966} & 1.74 & 8.225 \\
 & NeuS2        & \SecondBestCellColor{30.65} & 0.940 & 1.53  & 29.09 \\ 
 & VideoRF      & 29.22 & 0.883 & >20  & \SecondBestCellColor{0.554} \\
 & 3DGStream    & 27.34 & 0.856 & \BestCellColor{0.19} & 7.629  \\ \cline{2-6}
 & Ours         & \BestCellColor{32.28} & \SecondBestCellColor{0.946} & \SecondBestCellColor{0.89}  & \BestCellColor{0.513} \\ 
\hline 
\bottomrule
\end{tabular}
}
\end{center}
\caption{Quantitative comparison on ReRF~\cite{wang2023neural} dataset and HumanRF~\cite{icsik2023humanrf} dataset. Our method achieves the best rendering quality against other methods, achieving a fast training speed in a minute and a small storage of less than 600KB.}
\vspace {-7mm}
\label{tab:comparison}
\end{table}

\begin{figure*}[t]
    \centering
    \includegraphics[width=\linewidth]{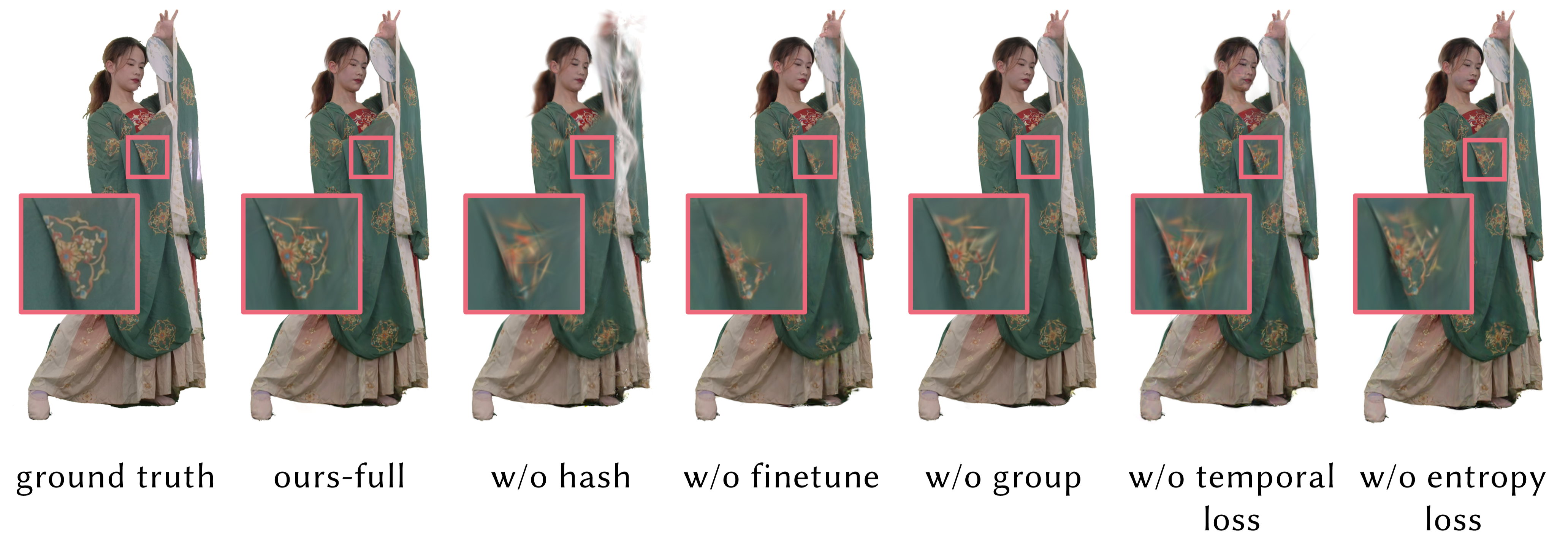}
    \caption{Qualitative evaluation of the performance of our various components at around 500KB, showing the necessity of each component in our methods. }
    \label{fig:ablation_fig}
\end{figure*}

\begin{figure}[t]
    \centering
    \vspace{-1mm}
    \includegraphics[width=\linewidth]{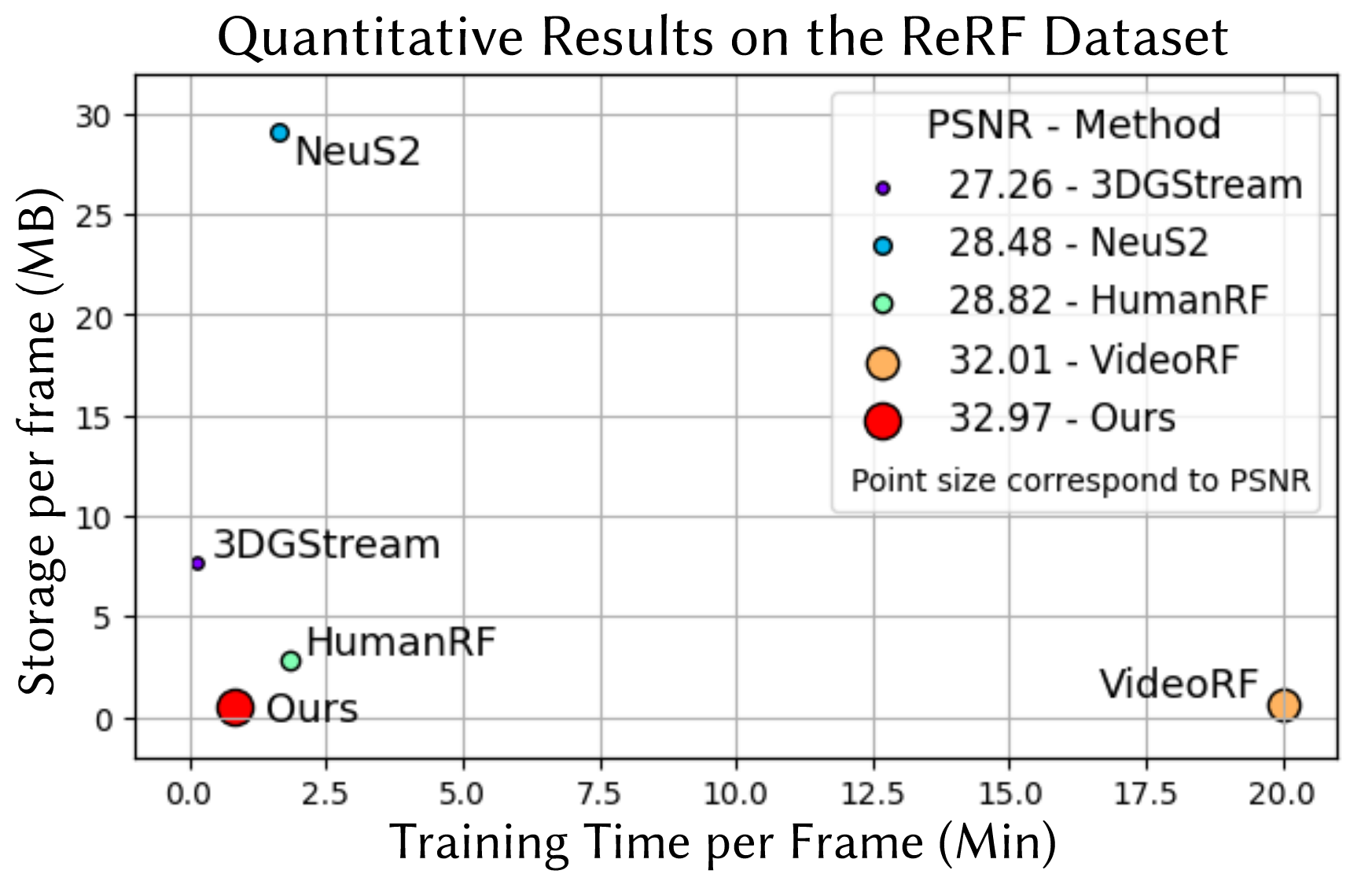}
    \vspace{-3mm}
    \caption{Compared with other methods, our method achieves fast training speed and high-quality rendering with small storage. }
    \vspace{-5mm}
    \label{fig:V}
\end{figure}

\subsection{Comparison} \label{sec:exp}
We compare V\textsuperscript{3} with the current state-of-the-art methods, including VideoRF~\cite{wang2024videorf}, 3DGStream~\cite{sun20243dgstream}, HumanRF~\cite{icsik2023humanrf}, and NeuS2~\cite{wang2023neus2}. These methods are evaluated on the Actors-HQ~\cite{icsik2023humanrf} dataset and the ReRF~\cite{wang2023neural} dataset for rendering quality, training time, and storage capacity. As illustrated in the Fig. ~\ref{fig:comparison}, VideoRF~\cite{wang2024videorf} and NeuS2~\cite{wang2023neus2}, which are based on neural voxel grids, resulting in a blurred effect in areas with fine textures. Although HumanRF~\cite{icsik2023humanrf} can represent a dynamic scene, it shows poor detail recovery. 3DGStream~\cite{sun20243dgstream} struggles with scenes involving large movements, as it relies primarily on the Neural Transformation Cache to obtain transformations between frames, yet it doesn't fine tune the Gaussians from the previous frame, making it challenging to handle complex sequences. Moreover, each frame requires querying an MLP to get the transformations, preventing efficient volumetric video streaming. In contrast, V\textsuperscript{3} achieves the most realistic human rendering effects with clearer texture details compared to other methods, while streaming in real-time.

We also conduct a quantitative comparison using metrics such as PSNR, SSIM, Training Time, and Storage Size. For the selection of comparison data, we chose the ReRF~\cite{wang2023neural} dataset, which contains large motion transformations, and the Actors-HQ~\cite{icsik2023humanrf} dataset. In the ReRF dataset, we follow the comparison method used in VideoRF~\cite{wang2024videorf} to evaluate the Kpop scene. For the Actors-HQ dataset, we use the test method described in the paper to evaluate Actor8, Sequence 1. 
As shown in Tab. ~\ref{tab:comparison} and Fig. ~\ref{fig:V}, our method outperforms others in both quality and storage across both datasets. We can ensure minimal quality degradation at lower storage capacities. In terms of training time, our method is second only to 3DGStream~\cite{sun20243dgstream}. Notably, compared to streamable methods like VideoRF~\cite{wang2024videorf} and 3DGStream~\cite{sun20243dgstream}, our method maintains high quality with smaller storage requirements. Additionally, the faster training speed of our method is more suitable for handling long sequences, making it more productive.

\begin{figure}[t]
    \centering
    \includegraphics[width=\linewidth]{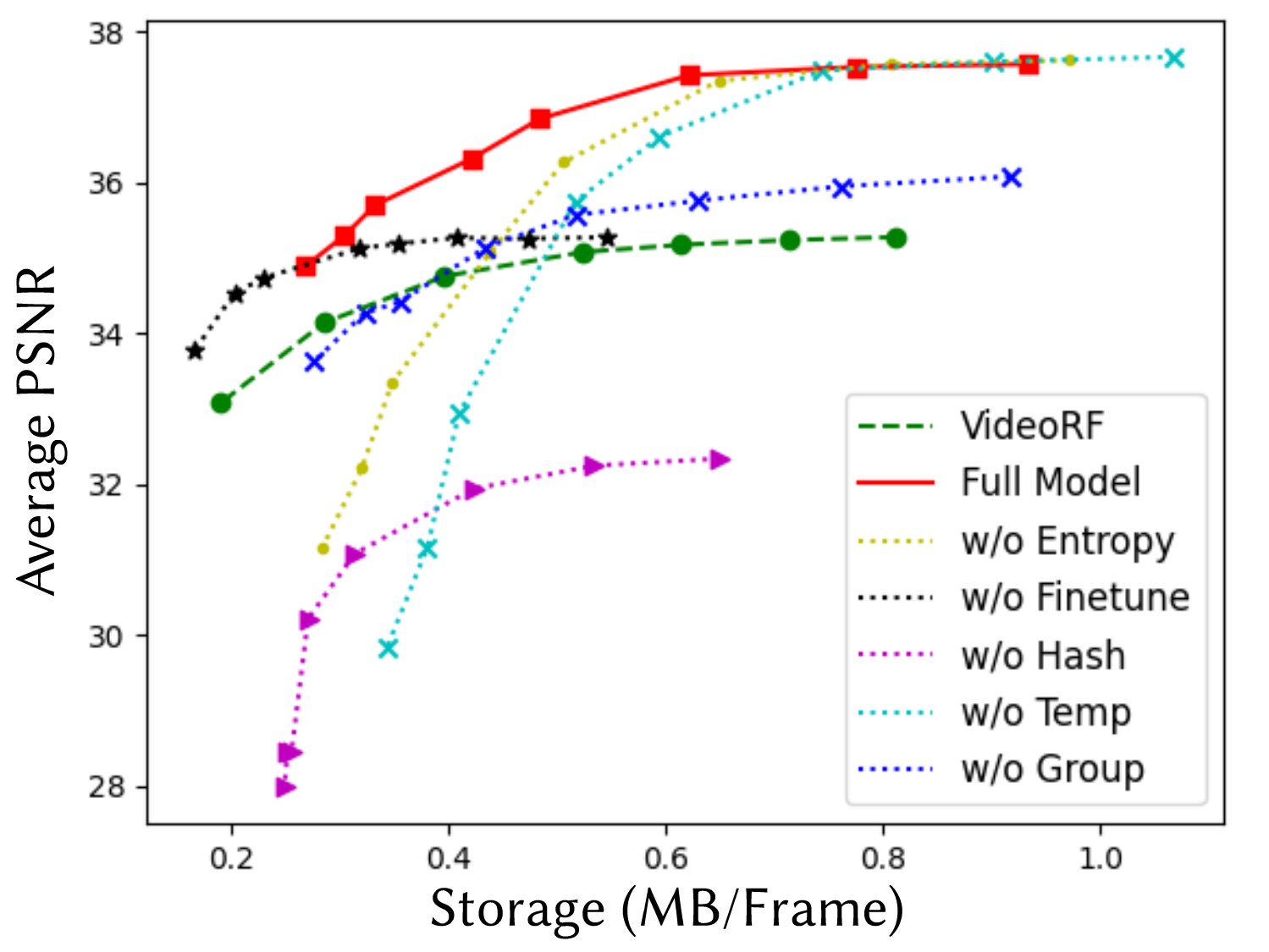}
    \caption{\textbf{Rate distortion Curve. } We test different components of our method under different codec compression QP settings. Our full model is at the top, allowing various storage needs while maintaining high-quality rendering. }
    \label{fig:distortion}
\end{figure}

\begin{figure*}[t]
    \centering
    \includegraphics[width=\linewidth]{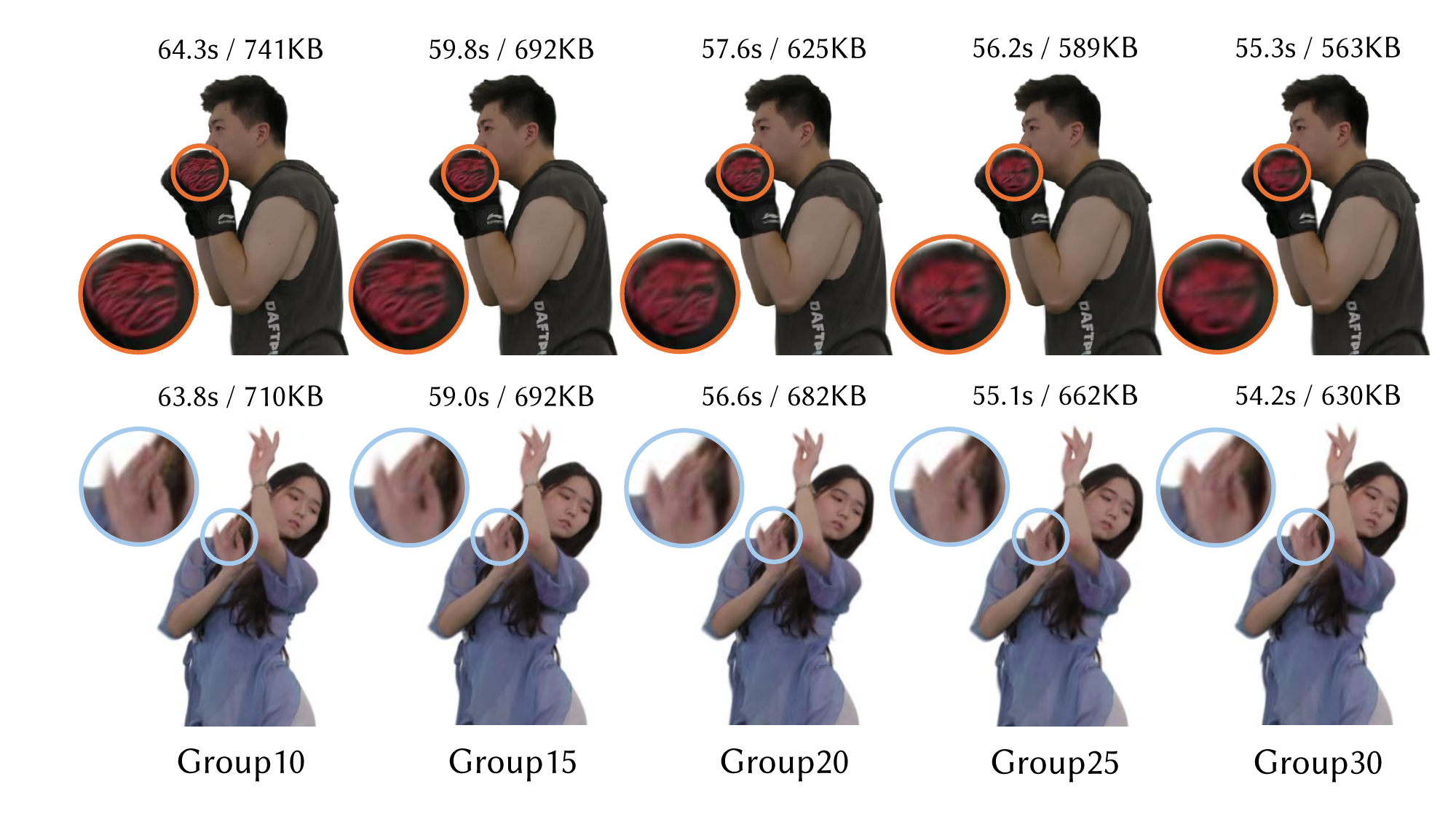}
    \caption{Qualitative evaluation of the grouping strategy indicates that the optimal number of frames per group is 20, which balances quality, storage requirements per frame, and training time per frame.}
    \label{fig:group_ablation}
\end{figure*}

\subsection{Evaluation} \label{sec:exp}
\paragraph{Ablation study.}
In this section, we further validate the effectiveness of the various components in our method. The qualitative and quantitative results are shown in Fig. ~\ref{fig:distortion}  and Fig. ~\ref{fig:ablation_fig}. Note that in Fig. ~\ref{fig:distortion}, all figures are obtained under storage conditions of approximately 600KB. We analyzed the impact of the residual entropy loss, temporal loss, hash-based motion estimation, fine tune, and 
frame group segmentation modules. The results indicate that without using the hash-based motion estimation, resulting in incorrect geometries and blurred appearances. After warping the point cloud, if no fine tuning is performed, certain details, such as clothing textures, will be challenging to reconstruct. Without the frame group segmentation strategy, the results of long-sequence training will suffer significant losses. Disabling temporal loss or residual entropy loss will result in video codec unfriendliness, leading to reduced quality under the same storage conditions.

\begin{table}[t]
\begin{center}
\small
\centering\setlength{\tabcolsep}{6pt}
\renewcommand{\arraystretch}{1.1}
\setlength{\tabcolsep}{1.5mm}{\begin{tabular}{c |c| ccc }
\hline
Platform  & FPS & Downloading & Decoding & Rendering \\ 
\hline
Desktop   & 435 & 7.8 ms & 13 ms  & 2.3 ms \\  
Tablet   & 96 & 13.2 ms & 10.9 ms & 10.4 ms \\ 
Phone  & 27 & 16.8 ms & 13.6 ms & 37.0 ms \\ 
\hline 
\bottomrule
\end{tabular}
}

\end{center}
\caption{Runtime analysis on multiple platforms of rendering under the resolution of $1920\times1080$ .}
\label{tab:devices}
\vspace{-4mm}
\end{table}




\paragraph{Multi-platform runtime analysis.}
We conducted a runtime analysis of our V\textsuperscript{3} player across multiple platforms. Our test platforms included an Ubuntu PC equipped with an Intel I9-10920X processor and an NVIDIA GeForce RTX 3090 GPU, an Apple iPad with an Apple M2 processor, and an Apple iPhone with an A15 Bionic processor. As shown in Tab. ~\ref{tab:devices},  we list the time consumption of each thread of the rendering pipeline. 
For the download thread, all platforms take about 10ms in an internal network setting. As for the decoding thread, the Desktop's multithreaded decoding combined with CUDA memory copying consumed 13ms. On the Apple mobile devices, the unified memory architecture eliminated the need for memory copying, and with parallel decoding using Compute Shaders, the time consumption was comparable to that of the Desktop.
As for rendering thread, the desktop equipped with a CUDA device achieves over 400 FPS, while mobile devices using metal can also render with a favorable FPS. As these three parts operate asynchronously and simultaneously, we use rendering time to calculate FPS. The well-designed asynchronous play pipeline ensures real-time volumetric video fetch, download, and rendering, guaranteeing an immersive volumetric video viewing experience across various devices.


\begin{table}[t]
\begin{center}
\small
\centering\setlength{\tabcolsep}{6pt}
\renewcommand{\arraystretch}{1.1}
\setlength{\tabcolsep}{1.5mm}{\begin{tabular}{l | c } 

\hline
        Stage & Time(s) \\
        \hline
            Average Frame Training & 56.1 \\
		Baking  & 0.5 \\ \hline
        Total Time & 56.6 \\
\hline 
\bottomrule
\end{tabular}
}
\end{center}
\caption{Runtime analysis of each stage in our training pipeline and post-baking. }
\label{tab:time}
\vspace{-7mm}
\end{table}

\paragraph{Training runtime analysis}
We also analyze the training time and baking time, shown in Tab. ~\ref{tab:time}. The key frame reconstruction including initial point cloud generation from NeuS2, 3DGS training, point cloud pruning, and fine tuning, totally takes 192.6s on average. The motion estimation will train for 500 iterations, taking 7.2s. In the fine tune stage, we will take an extra 2k iterations to optimize the Gaussian attributes, which takes 41.7s. In a frame group with 20 frames, each frame's reconstruction takes 56.1s on average. The process of baking including attribute quantization, Morton sorting, and using FFmpeg to perform H.264 encoding to 2D Gaussian videos, only takes 0.5s on average for each frame. As we set our frame group size to 20, our method can generate high-quality scenes in a minute for one frame, making the fast generation of volumetric video possible.


\begin{table}[t]
\begin{center}
\small
\centering\setlength{\tabcolsep}{6pt}
\renewcommand{\arraystretch}{1.1}
\setlength{\tabcolsep}{1.5mm}{\begin{tabular}{l | c | ccc }
\hline
 Dataset & Frame Number & PSNR & Training Time(s) & Size(KB) \\ 
\hline
\multirow{5}{*}{HiFi4G}
& 10 & 32.92  & 63.8 & 710  \\ \cline{2-5}
& 15 & 32.63  & 59.0 & 692  \\ \cline{2-5}
& 20 & 32.54  & 56.6 & 682  \\  \cline{2-5}
& 25 & 32.48  & 55.1 & 662  \\  \cline{2-5}
& 30 & 32.21  & 54.2 & 630  \\  \cline{2-5}
\hline
\bottomrule
\end{tabular}
}
\end{center}
\caption{Quantitative comparison of grouping strategy on HiFi4G~\cite{jiang2024hifi4g} dataset, where both training time and storage size are for each frame on average. }
\vspace {-7mm}
\label{tab:group tab}
\end{table}

\paragraph{Grouping strategy.}
To explore the optimal grouping strategy for $V^{3}$ training, we conducted an ablation study on the same dataset. Considering the efficiency of both training and inference, it is essential to balance training time and storage requirements when deciding on the grouping strategy.
As video codec achieves compression by reducing redundancy across multiple frames, so with the group length increases, the storage for each frame decreases. Additionally, since optimizing keyframes is time-consuming, larger groups can reduce the average training time per frame.
However, deformation between consecutive frames can accumulate errors, leading to a decline in the quality of frames positioned later in the group. Therefore, larger group sizes may result in decreased rendering quality. To explore the optimal group length, we conducted training with group sizes of 10, 15, 20, 25, and 30 frames, respectively.
As shown in Fig. ~\ref{fig:group_ablation} and Tab. ~\ref{tab:group tab}, our experimental results indicate that while storage requirements decrease with increasing frame group size, both training time and rendering quality also decrease. To achieve a balance between storage efficiency, training speed, and rendering quality, we chose a frame group size of 20 frames for our training.
\section{Limitations and Conclusion}
\paragraph{Limitation.}
Although we have proposed a method for streaming 2D Gridded Gaussians to mobile devices for viewing volumetric videos, our approach still has some limitations. First, since our method is based on Gaussians, achieving higher reconstruction quality requires dense camera views which is expensive, and poor segmentation when extracting foreground may result in some artifacts. Second, reconstructing large dynamic scenes, such as a grand stage performance with 30 participants or human-object interactions, may present challenges. Additionally, although we can reduce the training time to less than a minute per frame, we cannot achieve real-time reconstruction like GPS-Gaussian\cite{zheng2023gpsgaussian}, which could be a direction for future research.
\paragraph{Conclusion.}
We propose V\textsuperscript{3}, a novel method that enables the streaming and viewing of high-quality volumetric videos on mobile devices using 2D Gridded Gaussians. We innovatively bake 3D Gaussian attributes into 2D Gaussian video streams, leveraging video codecs for efficient compression before streaming to mobile devices for rendering. Additionally, we introduce an efficient training strategy that employs residual entropy loss and temporal loss to maintain high quality while ensuring temporal consistency, and enhancing video-codec friendliness. Our experiments demonstrate that we can achieve a compact, high-quality dynamic model with a relatively short training time. With this approach, we believe we can make a significant step towards the mobilization of volumetric videos, providing an unprecedented experience of streaming and viewing volumetric videos anytime and anywhere, with seamless video scrolling and sharing capabilities.

\begin{acks}
This work was supported by the National Key R\&D Program of China (2022YFF0902301), NSFC programs (61976138, 61977047), STCSM (2015F0203-000-06), and SHMEC(2019-01-07-00-01-E00003). We also acknowledge support from the Shanghai Frontiers Science Center of Human-centered Artificial Intelligence and MoE Key Lab of Intelligent Perception and Human-Machine Collaboration (ShanghaiTech University).
\end{acks}

\bibliographystyle{ACM-Reference-Format}
\bibliography{references.bib}

\end{document}